# Sentiment Dynamics of Success: Fractal Scaling of Story Arcs Predicts Reader Preferences

Yuri Bizzoni
au701203@uni.au.dk

Telma Peura
tpeura@cc.au.dk

Mads Rosendahl Thomsen
madsrt@cc.au.dk

Kristoffer Nielbo
kln@cas.au.dk

## Abstract

We explore the correlation between the sentiment arcs of H. C. Andersen's fairy tales and their popularity, measured as their average score on the platform GoodReads. Specifically, we do not conceive a story's overall sentimental trend as predictive per se, but we focus on its coherence and predictability over time as represented by the arc's Hurst exponent. We find that degrading Hurst values tend to imply degrading quality scores, while a Hurst exponent between .55 and .65 might indicate a "sweet spot" for literary appreciation.

## 1 Introduction and Related Work

What determines the perception of literary quality? It can be argued that the feelings expressed by a story play a major role in its experience and assessment, since they activate similar sentiments in the readers' memory (Drobot, 2013; Hu et al., 2021b). At the same time, the relationship between a reader and the sentiments expressed by a text is anything but linear: the reader's perception of the sentiments on page depends on their context (the victory of a villain can engender unpleasant feelings, Maks and Vossen (2013)), and mechanisms like irony and catharsis can transform the negative feelings expressed by a text into positive feelings for a reader (Bosco et al., 2013). Ultimately, the intensity and nature of feelings on page does not tell us much about whether a reader will love, hate or remain indifferent towards a story. While it is of great interest to explore the temporal dimension of sentiments in literary texts (Gao et al., 2016; Cambria et al., 2017), and specific tools have been developed to this aim (Brooke et al., 2015; Jockers, 2017), it might not be the explicit sentimental trend of a story which plays a role in its quality, but the distribution and relative change of such sentiments through the story, independently from whether the direction of the narrative's arc is closer to a comedy or to a tragedy. We use nonlinear adapting filtering and fractal analysis to assess the inner coherence of a story arc and the auto-correlation in the sentiment dynamics at different time-scales, as measured through the Hurst exponent (Hu et al., 2021b). We then explore the correlation between such value and perceived literary quality.

We propose H. C. Andersen's fairy tales as an ideal test case for our hypothesis: Andersen's tales tend to have relatively simple narratives, making it easier for us to explain their arcs and their level of predictability; they are well known and widely read in contemporary English, allowing us to retrieve the average rating from many readers (in the most popular cases, tens of thousands of readers); and to use state-of-the-art sentiment analysis tools. Lastly, some stories are short, which is a disadvantage when producing sentiment arcs, but is also a significant advantage when interpreting our results since we can re-read full stories in a short time and be assured of the quality of the computed arcs, making it much easier to explore the underlying causes of sentimental curves and their resulting Hurst exponents. While deciding whether a story's segment is happy or sad is a complex matter of interpretation, we can measure the sentimental value of the components from which such interpretations derive: the average positive or negative value of its words or sentences.

During the last two decades, sentiment analysis has generated a large number of resources to infer the kind and intensity of the sentiment expressed by a text, at the word (Mohammad and Turney, 2013), phrase (Agarwal et al., 2009; Hutto and Gilbert, 2014), sentence (Hu and Liu, 2001) or overall text level (Pang and Lee, 2004). Word-level resources, which are mainly constituted of manually or semi-automatically annotated lexica (Taboada et al., 2011), are popular in literary senti-ment analysis (Gao et al., 2016). They provide the highest number of data points for each story, and they keep the inference of the arcs at the simplest possible level, requiring neither rule-based nor pre-trained scoring systems. While this is a limitation in the endeavour of inferring the sentiment of a portion of text, since sentiment lexica do not disam-biguate the sense of a word in context (Zhang and Liu, 2017) and the analysis is limited to the words present in the used lexicon, this approach allows for drawing the succession of the smallest sentimental units of a text, from which human readers them-selves draw their inferences. Furthermore, using lexica allows for drawing arcs in a transparent way, making it possible for researchers to immediately understand the causes of each score derived from a corpus. Finally, we use word-level sentiment arcs and their large-scale human ratings to determine whether the dynamic sentiment evolution and the level of coherence of a story arc at the sentiment level are connected with the perceived quality of the stories.

## 2 Data

Andersen corpus Our corpus consists of a col-lection of 126 H. C. Andersen's fairy tales trans-lated into English retrieved from Project Guten-berg[1]. Their length varies between 1956 characters (The Princess and the Pea) and 106496 characters (The Ice Maiden), with an average length of 1585 characters.

SA lexicon To create the stories' sentiment arcs, we rely on the NRC-VAD lexicon (Mohammad, 2018), which is composed of almost 20.000 En-glish words annotated for valence, arousal and dom-inance [2]. For this study, we only used valence.

GoodReads scores To measure the stories' per-ceived quality we resorted to GoodReads (Thelwall and Kousha, 2017), a popular web platform used to grade, comment and recommend books. We manu-ally collected the average rating and the number of individual raters for each of Andersen's fairy tales (see Figure 1).

## 3 Methods

We proceed in three main steps: (i) we convert the stories into raw sentimental arcs through the use of the NRC-VAD lexicon (Section 3.1); (ii) we compute the story arcs' inner coherence through their Hurst coefficient (Section 3.2); (iii) we corre-late each story's Hurst coefficient with its average rating on GoodReads (Section 4).

### 3.1 Arc Extraction

By retrieving the sentiment value of each word in our lexicon, we created a fine-grained sentiment arc for each story in the dataset. Since the lexicon we used only contains words with a non-neutral value, we assigned all out-of-vocabulary words a neutral value by default, in order to represent not only the highs and lows but the neutral sequences of the stories as well. After creating the arcs, we applied a smoothing technique to both improve their visualization and facilitate their detrending in post processing (see Figure 2). To control for the sensibility of our arcs, we selected a subset of 12 popular Andersen stories and grouped them in hierarchical sets through agglomerative clustering (Murtagh and Legendre, 2014). Two of the au-thors then independently checked the main clusters, verifying that they represented major, generally co-herent sentimental groups.

### 3.2 Arc Coherence

Following Hu et. al (Hu et al., 2021a,b), we use the Hurst exponent to approximate the story arc's inner coherence. The Hurst exponent, H, is a mea-sure of self-similar behavior. In the context of story arcs, self-similarity means that the arc's fluctuation patterns at faster time-scales resemble fluctuation patterns at slower time scales (Riley et al., 2012). We use Adaptive Fractal Analysis (AFA) to esti-mate the Hurst exponent (Gao et al., 2011).

AFA is based on a nonlinear adaptive multi-scale decomposition algorithm (Gao et al., 2011). The first step of the algorithm involves partitioning an arbitrary time series under study into overlapping segments of length w = 2n + 1, where neighbor-ing segments overlap by n + 1 points. In each segment, the time series is fitted with the best poly-nomial of order M, obtained by using the standard least-squares regression; the fitted polynomials in overlapped regions are then combined to yield a single global smooth trend. Denoting the fitted polynomials for the i − th and (i + 1) − th seg-ments by $y^i(l_1)$ and $y^{(i+1)}(l_2)$, respectively, where l1, l2 = 1, · · · , 2n + 1, we define the fitting for the

---
[1] https://www.gutenberg.org/ebooks/27200
[2] We also experimented with the Vader lexicon (Hutto and Gilbert, 2014), but found its coverage too small to yield mean-ingful results at the word level.

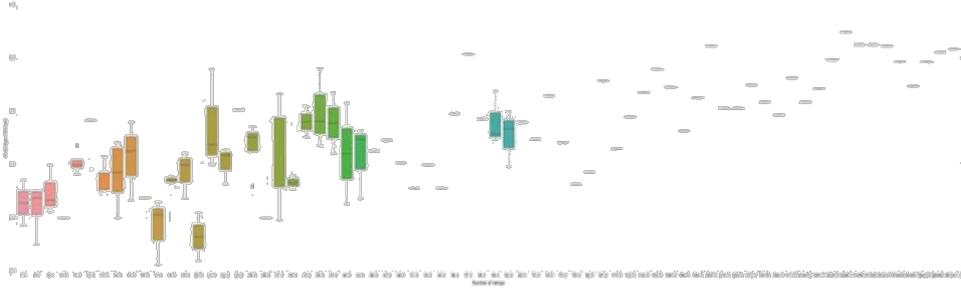

Figure 1: Number and magnitude of ratings. Most stories have less than 100 ratings, but the most known tales reach up to 40 thousand individual scores; stories with more raters tend to also receive higher average scores.

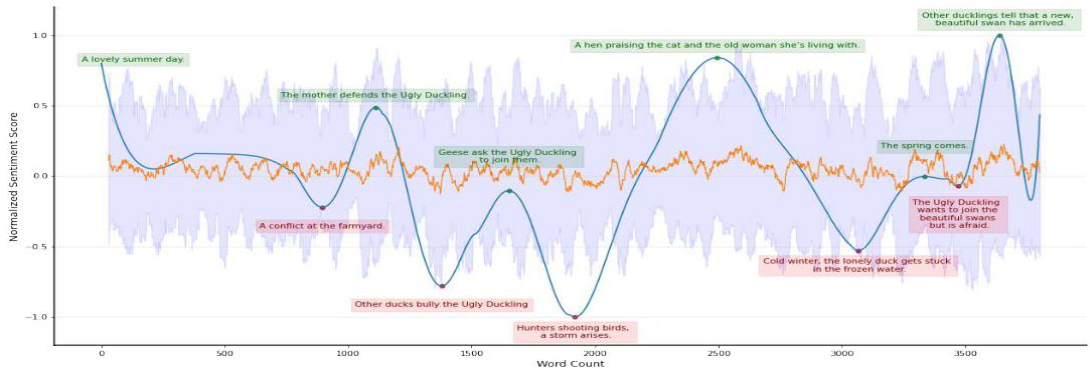

Figure 2: Sentiment arc from The Ugly Duckling with its narrative references. In orange and lilac, the mean and standard deviation of sentiment scores for every 30 words; the blue line is the polynomial smoothing of the raw sentiment arc.

overlapped region as

$$y^{(c)}(l) = w_1 y^{(i)}(l+n) + w_2 y^{(i+1)}(l),$$
$$l = 1, 2, \cdots, n+1$$

where $w_1 = \left(1 - \frac{l-1}{n}\right)$ and $w_2 = \frac{l-1}{n}$ can be written as $(1-d_j/n)$ for $j = 1, 2$, and where $d_j$ denotes the distances between the point and the centers of $y^{(i)}$ and $y^{(i+1)}$, respectively. Note that the weights decrease linearly with the distance between the point and the center of the segment. Such a weighting is used to ensure symmetry and effectively eliminate any jumps or discontinuities around the boundaries of neighboring segments. As a result, the global trend is smooth at the non-boundary points, and has the right and left derivatives at the boundary (Riley et al., 2012). The global trend thus determined can be used to maximally suppress the effect of complex nonlinear trends on the scaling analysis. The parameters of each local fit is determined by maximizing the goodness of fit in each segment. The different polynomials in overlapped parts of each segment are combined so that the global fit will be the best (smoothest) fit of the overall time series. Note that, even if M = 1 is selected, i.e., the local fits are linear, the global trend signal will still be nonlinear. With the above procedure, AFA can be readily described. For an arbitrary window size w, we determine, for the random walk process u(i), a global trend v(i), i = 1, 2, · · · , N, where N is the length of the walk. The residual of the fit, u(i) − v(i), characterizes fluctuations around the global trend, and its variance yields the Hurst parameter H according to the following scaling equation:

$$F(w) = \left[\frac{1}{N}\sum_{i=1}^{N}(u(i) - v(i))^2\right]^{1/2} \sim w^H$$

By computing the global fits, the residual, and the variance between original random walk process and the fitted trend for each window size w, we can plot log2 F (w) as a function of log2 w. The pres-ence of fractal scaling amounts to a linear relation in the plot, with the slope of the relation provid-ing an estimate of $H$[3]. Accordingly, a H higher than .5 indicates a degree of linear coherence (e.g., positive sentiments are followed by positive sen-timents), while H lower than .5 indicates a series that tends to revert to the mean (e.g. a positive emotion always follows a negative emotion). Sto-ries based on the repetition of the same narrative mechanism like T heButterfly, where a butterfly meets several nice flowers but always finds them faulty, have a relatively low Hurst exponent (see Figure 4).

---

[3]Code for computing DFA and AFA is available at https://github.com/knielbo/saffine.

## 4 Results

Our final step consisted in correlating each fairy tale's average rating with its average Hurst coefficient. We found that there is a small, but significant correlation between the Hurst exponent of Ander-sen's tales and their average success on Good Reads (see Figure 3).

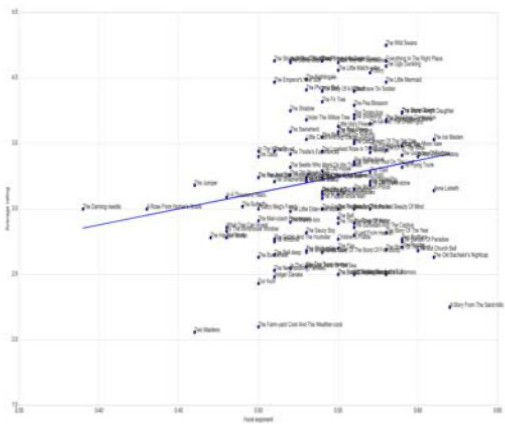

Figure 3: Correlation between Hurst exponent and aver-age rating for all tales.

Nonetheless, there appears to be a difference in such correlation when we consider the number of ratings each story received. As we discussed before, the statistical value of those ratings can change: ob-scure stories can have less than ten different raters, while the most known fairy tales can be rated by tens of thousands of readers. Many of Andersen's stories are not well known to the larger public, and thus received a relatively small number of reviews. Based on the average number of ratings of each story, we recomputed the correlation strength for only those tales that received more than 30 different scores, which excluded around half of our corpus, leaving 63 tales in the corpus. This threshold al-lowed us to keep only stories that had a significant amount of individual annotations, without excessively reducing the size of our dataset. The result was a much stronger correlation, and a more significant p-value (see Figure 4). In Table 1, we show a summary of the correlation values we obtained on both sets. Finally, in Figure 5 we draw the overall intuition of the study: works with a smaller Hurst exponent feature mean-reverting sentiment trends and elicit lower overall appreciation.

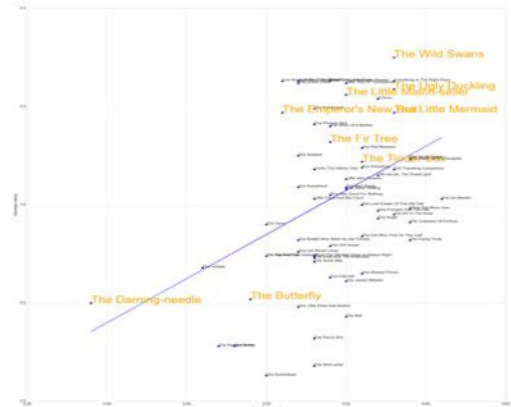

Figure 4: Correlation between Hurst exponent and average rating for tales having more than 30 ratings with some titles in evidence. Most of the very popular stories fall on the upper right corner, while tales like The But-terfly, based on a simple repetition of the same dynamic, fall more to the left. Also notice how without the outlier The Darning Needle, the correlation of the remaining data points would be even steeper.

|  | All tales | | Popular tales | |
|---|---|---|---|---|
|  | corr. | p value | corr. | p value |
| Pearson | .19 | .03 | .4 | .001 |
| Spearmann | .18 | .04 | .35 | .005 |
| Kendall Tau | .12 | .03 | .23 | .009 |
| Distance corr. | .81 | | .6 | |

Table 1: Correlations for all tales and tales with more than 30 ratings ("popular"). Statistical significance is not directly applicable to standard distance correlation.

## 5 Conclusions and Future Work

The main finding of this paper is that there is a correlation between a story's sentimental coherence and its perceived quality. With all the necessary caveats, we find that such correlation is an interesting result and advocates for a more extensive use of multifractal theory in the study of sentimental arcs in literature. The emotional coherence of a story, as represented by the average Hurst exponent of the sentiment arc, explains a part of its perceived quality as measured by its average rating on a gen-eral audience website. Furthermore, stories with more ratings tend to show a stronger correlation between these two measures, which might mean that the weaker correlations we recorded for the less known stories might be due to an insufficiently large pool of raters (thus having a less robust score). It is interesting to notice that stories with many ratings also tend to have higher ratings on aver-age: this further shows how, at least for fairy tales, fame and popularity tend to go together. While it proved surprisingly predictive, reducing a story arc to one overall Hurst coefficient means losing important information in terms of how coherence is distributed through the narrative. Exploring the temporal variation of this coefficient at different time scales of a story might reveal further insights into the sentiment dynamics of a good narrative evolution. Another aspect for further exploration is evaluating alternative ways of scoring the senti-ments in a text. Particularly, when applied to longer stories, adopting a sentence-level approach could help accounting for the context that indeed affects the sentiment interpretation. Furthermore, instead of sentiment analysis, moving to an emotion analy-sis might allow for more detailed insights about an optimal narrative development. However, already the present results suggest that there exists a desir- able ratio of coherence and unpredictability in the sentiment arcs that contribute to the appreciation of a story.

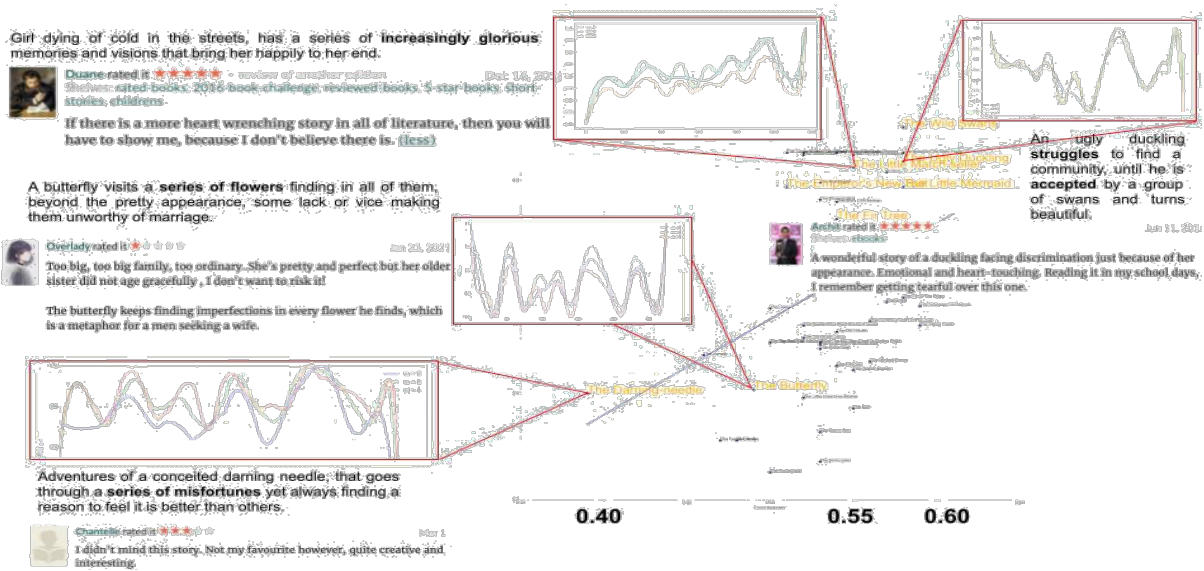

Figure 5: A visual summary of our concept. More "zig-zag" lines reverting to the mean tend to receive less favorable overall reviews than stories having a smoother trendline. The latter ones also include many of the most known Andersen's stories.

## Acknowledgments

This paper has been supported the 'Fabula-NET: A Deep Neural Network for Automated Multidi-mensional Assessment of Literary Fiction and Nar-ratives' funded by the Velux Foundation, and the DeiC Interactive HPC system with project DeiC-AU1-L-000015.